\documentclass[10pt,twocolumn,letterpaper]{article}

\usepackage[pagenumbers]{cvpr} 
\usepackage[normalem]{ulem}
\usepackage{multirow}
\usepackage[table,xcdraw]{xcolor}
\useunder{\uline}{\ul}{}
\usepackage{comment}

\definecolor{cvprblue}{rgb}{0.21,0.49,0.74}
\usepackage[pagebackref,breaklinks,colorlinks,allcolors=cvprblue]{hyperref}

\title{Explaining Without Forgetting: Bridging Off-the-Shelf Detectors with a Multimodal LLM for Explainable Deepfake Detection}

\author{Evlampios Apostolidis\\
ITI - CERTH\\
Thessaloniki, Greece\\
{\tt\small apostolid@iti.gr}
\and
Dimitrios Kaltsios\\
ITI - CERTH\\
Thessaloniki, Greece\\
{\tt\small dkaltsios@iti.gr}
\and
Vasileios Mezaris\\
ITI - CERTH\\
Thessaloniki, Greece\\
{\tt\small bmezaris@iti.gr}
}

\begin{document}
\maketitle
\providecommand{\ewf}{EwF}

\begin{abstract}
State-of-the-art deepfake detectors can be highly accurate but operate as black boxes, and their decisions are typically explained through saliency maps that show where a detector focuses but not why, leaving the interpretation to human experts. To explain decisions in natural language while retaining the forensic capability of such detectors, we propose \ewf{} (Explaining without Forgetting), a detector-agnostic framework in which a frozen, off-the-shelf deepfake detector acts as a forensic expert for a multimodal large language model (MLLM). A lightweight adapter maps the detector's spatial feature map, location by location, into forensic tokens that enter the MLLM next to its native image tokens, so that a new detector can be plugged in by training only the adapter and performing low-rank adaptation of the MLLM. A staged training scheme first adapts the framework to deepfake detection, on data that match the detector's training regime, and then teaches it to explain its decisions. We couple Qwen3-VL-4B with four detectors of different architectures and training regimes, and evaluate intra-dataset (FaceForensics++, two compression levels) and cross-dataset (Celeb-DF, DFDC, WildDeepfake) detection performance, as well as detection and explanation performance on DD-VQA. Paired statistical tests show that \ewf{} preserves or significantly improves the detection performance of every coupled detector, while its explanations outperform those of state-of-the-art methods for explainable deepfake detection. Code will be made publicly available upon acceptance.
\end{abstract}    
\providecommand{\ewf}{EwF}

\section{Introduction}
\label{sec:intro}

The rapid progress of generative models has made it possible to create realistic videos where one or more faces have been manipulated; these videos are commonly known as deepfakes. Such content can be used to spread disinformation, damage reputations, and enable fraud and identity theft~\cite{TOLOSANA2020131}. To tackle this challenge, considerable effort has been devoted to the development of deepfake detectors based on deep learning architectures. State-of-the-art detectors (\eg, \cite{9880195,10657482}) exhibit high accuracy on established benchmarks~\cite{9010912,9156368,DFDC2020,10.1145/3394171.3413769}, but operate as black boxes that output only a binary decision along with a probability score. In applications such as journalism, content moderation and digital forensics, this is hardly sufficient: users need to understand \emph{why} a video is deemed manipulated in order to trust, verify and act upon the decision.

\begin{figure}[t]
  \centering
  \includegraphics[width=\columnwidth]{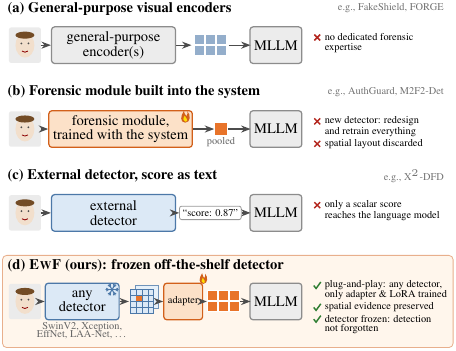}
  \caption{How forensic evidence reaches the language model in explainable deepfake detection: (a) general-purpose visual encoders~\cite{ICLR2025_4d4e0ab9,kundu-etal-2026-forge}; (b) a forensic module designed and trained as part of the system~\cite{11492581,11092358,sun2025edvdllama}; (c) the output score of an external detector, inserted as text~\cite{NEURIPS2025_78a2ed8e}; (d) \ewf{}: the full spatial feature map of a frozen detector, mapped into per-location forensic tokens.}
  \label{fig:teaser}
\end{figure}

Early work on explainable deepfake detection produced visual explanations~\cite{10.1145/3699710}, as the available explanation tools were designed to attribute the decision of an image classifier to input regions. Such explanations included saliency maps from post-hoc methods (\eg, \cite{10.1145/2939672.2939778,10.5555/3295222.3295230,8354201}) indicating the facial regions that most influenced a detector's decision (\eg, \cite{9092227,DBLP:journals/corr/abs-2105-05902,9993294,SILVA2022100217,10972511}), attention or manipulation maps produced by the detector itself~\cite{9156378,9157215}, and prototypes of artifact patterns~\cite{9423253}. However, such explanations show where the detector looks but leave the reasons -- which artifact, and why it betrays a manipulation -- to be inferred by a human expert. This limitation, together with the advent of multimodal large language models (MLLMs), motivated a shift towards explanations in natural language, which are easier to understand for non-experts and more informative for experts. An early study prompted general-purpose MLLMs, such as GPT-4V, to detect deepfakes~\cite{Jia_2024_CVPR}, while Zhang \etal~\cite{10.1007/978-3-031-73223-2_22} formulated explainable deepfake detection as a visual question answering task and introduced the DD-VQA dataset with human-written explanations about whole face images and individual facial regions. Subsequent methods fine-tuned various MLLMs to produce natural language explanations~\cite{11492581,11092358,ICLR2025_4d4e0ab9,sun2025edvdllama,NEURIPS2025_78a2ed8e,kundu-etal-2026-forge}.

In most of these methods (Fig.~\ref{fig:teaser}), the forensic component is either a general-purpose visual encoder~\cite{ICLR2025_4d4e0ab9,Jia_2024_CVPR,10.1007/978-3-031-73223-2_22,kundu-etal-2026-forge} or a module designed and trained as an integral part of the system~\cite{11492581,11092358,sun2025edvdllama}. X2-DFD~\cite{NEURIPS2025_78a2ed8e} does exploit external detectors, but only through their output scores, inserted into the prompt as text. Hence, existing methods cannot fully benefit from progress in dedicated deepfake detectors: adopting a stronger detector architecture either entails redesigning the fusion mechanism and retraining the whole architecture, or limits its contribution to conveying only its final decision score. Moreover, the forensic evidence is often condensed into a single pooled embedding~\cite{11092358} or serialized into text~\cite{ICLR2025_4d4e0ab9,sun2025edvdllama,NEURIPS2025_78a2ed8e}, discarding the spatial layout that could ground an explanation on a specific facial region. Finally, these methods are compared, at best, with published detectors resembling their forensic component, rather than with the integrated detector itself under the same protocol; thus, whether learning to explain preserves the accuracy of the very detector being integrated, in particular on unseen data, remains unexamined.

To address these limitations, we propose \ewf{} (Explaining without Forgetting), a framework that treats an off-the-shelf deepfake detector as a frozen forensic expert and couples it with an MLLM for explainable deepfake detection. The detector's final spatial feature map is mapped location by location into forensic tokens that enter the MLLM next to its native image tokens. Any detector exposing such a map can be plugged in by training only a lightweight adapter and performing a low-rank adaptation (LoRA)~\cite{hu2022lora} of the MLLM, while the detector remains untouched, so the detection capability it brings is not forgotten. Our contributions are as follows:
\begin{itemize}
    \item \textbf{Plug-and-play forensic-to-multimodal alignment.} A detector-agnostic framework in which a frozen deepfake detector acts as a forensic expert for an MLLM through its full spatial feature map, mapped into per-location forensic tokens that preserve the spatial layout of the evidence; new detectors are incorporated by training only an MLP adapter and the MLLM's LoRA weights.

    \item \textbf{Forensic adaptation before explanation.} A staged training scheme that adapts the MLLM to real/fake classification on data that match the detector’s training regime (FF++ forgeries or self-blended images), then aligns the adapter and performs another round of low-rank adaptation of the MLLM on the same data, and only then fine-tunes the framework for explanation generation, so that learning to explain preserves the detection capability.

    \item \textbf{Extensive evaluation.} We instantiate \ewf{} with Qwen3-VL-4B~\cite{Qwen3-VL} and detectors of different architectures and training regimes (SwinV2, XceptionNet and EfficientNet-B4, trained on real and manipulated FF++ data, and LAA-Net~\cite{10657482}, trained on self-blended images), and evaluate its detection performance intra-dataset (FF++, at two compression levels) and cross-dataset (Celeb-DF, DFDC, WildDeepfake), with paired statistical tests against each detector. Moreover, we assess its detection and explanation performance on DD-VQA at both compression levels. Our findings indicate that \ewf{}, aligned on data that match the detector's training regime, preserves or significantly improves the detection performance of every coupled detector, while producing explanations that outperform those of state-of-the-art explainable methods.
\end{itemize}

\providecommand{\ewf}{EwF}

\section{Related work}
\label{sec:related_work}

Early work aimed to explain the decisions of deepfake detectors visually~\cite{10.1145/3699710}, through saliency maps produced by post-hoc methods~\cite{9092227,DBLP:journals/corr/abs-2105-05902,9993294,SILVA2022100217,10972511}, learned attention or manipulation maps~\cite{9156378,9157215}, prototypes of artifact patterns~\cite{9423253}, or disentangled source and target artifacts~\cite{10.1007/978-3-031-19781-9_2}. As visual explanations leave the interpretation of the highlighted regions to a human expert, research has shifted towards explanations in natural language. Jia \etal~\cite{Jia_2024_CVPR} were among the first to study whether general-purpose MLLMs can detect deepfakes without any training. They prompted GPT-4V and Gemini to distinguish real from synthetic face images and justify their decisions, and showed that the models relied mostly on semantic-level cues, were quite sensitive to the prompt, and remained less accurate than recent dedicated detectors. Zhang \etal~\cite{10.1007/978-3-031-73223-2_22} cast explainable deepfake detection as visual question answering and fine-tuned BLIP with additional text and image contrastive objectives (BLIP-TI) on face- and region-level question--answer pairs; the learned representations can be fused into existing detectors, but no forensic features reach the language model. 

\begin{figure*}[t]
  \centering
  \includegraphics[width=0.97\textwidth]{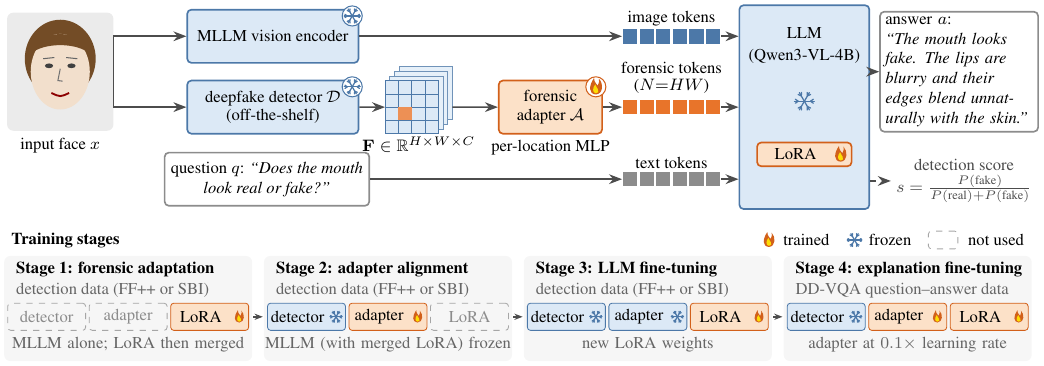}
  \caption{Overview of \ewf{}. Top: a frozen, off-the-shelf deepfake detector provides a spatial feature map; a per-location adapter maps every location to a forensic token in the embedding space of the language model; the MLLM (Qwen3-VL-4B, whose LLM is shown on the right) reads the image tokens, the forensic tokens and the question, and answers with a verdict and an explanation, from which a continuous detection score is also read. Bottom: the four training stages, with the modules trained and frozen in each.}
  \label{fig:overview}
\end{figure*}

More recent methods fine-tune MLLMs and differ mainly in how forensic evidence reaches the language model. A first group relies on general-purpose visual representations. FORGE~\cite{kundu-etal-2026-forge} complements the contrastive encoder of the MLLM (PaliGemma2) with the patch tokens of a self-supervised encoder (DINOv2), which are mapped by a lightweight adapter and interleaved with the original visual tokens, and fine-tunes the language model on face-manipulated and fully synthetic images. FakeShield~\cite{ICLR2025_4d4e0ab9} addresses general image forgery detection and localization, including deepfakes, by conditioning a LLaVA-based model on a textual domain tag predicted by an auxiliary classifier and decoding a manipulation mask with SAM~\cite{Kirillov_2023_ICCV}. A second group introduces forgery-specific visual representations. AuthGuard~\cite{11492581} trains a CLIP vision encoder with a binary classification loss and an image--text contrastive loss on pseudo-captions generated by a large language model, and then connects it to a LLaVA-style MLLM. M2F2-Det~\cite{11092358} combines a CLIP encoder, adapted through forgery prompt learning, with a dedicated deepfake encoder; a bridge adapter fuses the two streams into a pooled forgery embedding that is passed to the language model together with the visual tokens. EDVD-LLaMA~\cite{sun2025edvdllama} extends the setting to videos, fusing a spatio-temporal Swin Transformer branch over grids of consecutive frames with SigLIP features, and provides the language model with hand-crafted facial integrity measurements serialized as text, within a chain-of-thought reasoning framework. Closest to our work, X2-DFD~\cite{NEURIPS2025_78a2ed8e} couples a LLaVA-based MLLM with dedicated detectors. It first assesses which forgery features the MLLM can already recognize, fine-tunes it on explanations generated by GPT-4o around these strong features, and supplements the weak ones with external detectors, such as the blending-based CDFA detector~\cite{10.1007/978-3-031-73016-0_7}, whose output probability is inserted as text into the prompt. This makes the framework extendable with new detectors, but each detector contributes only a scalar score instead of its feature representation.

In all these approaches, the forensic component is either absent, replaced by a general-purpose encoder, designed and trained as an integral part of the proposed system, or reduced to the scalar output of an external detector conveyed as text. In contrast, \ewf{} treats an existing, independently trained deepfake detector as a frozen forensic expert and passes its full spatial feature map into the embedding space of the MLLM, to the language model. Any detector exposing such a map can be plugged in without modification, and the spatial layout of the forensic evidence is preserved for region-level explanations.

\providecommand{\ewf}{EwF}

\section{Proposed method}
\label{sec:method}

Given a face image $x$ and a natural-language question $q$ about its authenticity (\eg, whether the face, or a facial region, looks real or fake), \ewf{} generates an answer $a$ that contains both a verdict and an explanation. As shown in Fig.~\ref{fig:overview}, \ewf{} consists of three components: (i)~an off-the-shelf deepfake detector $\mathcal{D}$, trained independently and kept frozen; (ii)~a lightweight forensic adapter $\mathcal{A}$, which maps the detector's spatial feature map, location by location, into the token-embedding space of the language model; and (iii)~an MLLM, which receives the image and the forensic tokens, and generates the answer. The adapter is the only detector-specific module, so a new detector is incorporated by training a new adapter and performing low-rank adaptation of the MLLM.

\subsection{Forensic tokens from a frozen detector}
\label{sec:method_tokens}

Let $x_{\mathcal{D}}$ denote the image $x$ preprocessed as in the detector's own training (crop, input size and normalization); the MLLM receives $x$ through its own image processor. Let $\mathcal{D}_{\text{feat}}$ denote the part of the detector up to its last spatial feature map, before global pooling and classification,
\begin{equation}
  \mathbf{F} = \mathcal{D}_{\text{feat}}(x_{\mathcal{D}}) \in \mathbb{R}^{H \times W \times C},
\end{equation}
where $H$ and $W$ are the spatial dimensions and $C$ the number of channels. We flatten $\mathbf{F}$ in raster order into $N = HW$ vectors $\mathbf{f}_1,\dots,\mathbf{f}_N \in \mathbb{R}^{C}$. No pooling is applied, so the spatial layout of the forensic evidence is preserved and each token corresponds to a facial area that an explanation can refer to. The detector runs in inference mode and without gradients throughout. For detectors whose decision heads do not read the last backbone stage, the map is taken where the heads read it and pooled to a fixed grid (for LAA-Net~\cite{10657482}, the fused E-FPN map, average-pooled to $12 \times 12$).

The adapter $\mathcal{A}$ is a two-layer MLP shared across locations, which maps each $\mathbf{f}_i$ to a forensic token $\mathbf{t}_i \in \mathbb{R}^{d}$,
\begin{equation}
  \mathbf{t}_i = \mathbf{W}_2\,\mathrm{GELU}\big(\mathbf{W}_1\,\mathrm{LN}(\mathbf{f}_i) + \mathbf{b}_1\big) + \mathbf{b}_2,
  \quad i = 1,\dots,N,
\end{equation}
where $d$ is the hidden size of the language model, layer normalization ($\mathrm{LN}$) brings the detector-specific feature statistics to a common scale, GELU is the Gaussian error linear unit~\cite{hendrycks2016gelu}, and $\mathbf{W}_1 \in \mathbb{R}^{h \times C}$, $\mathbf{b}_1 \in \mathbb{R}^{h}$, $\mathbf{W}_2 \in \mathbb{R}^{d \times h}$ and $\mathbf{b}_2 \in \mathbb{R}^{d}$ are learned, with a hidden size $h$. Only $C$ and $N$ depend on the detector.

\subsection{Injecting forensic tokens into the MLLM}
\label{sec:method_injection}

We add a placeholder token \texttt{<forensic>} to the vocabulary and form the input sequence of the MLLM as
\begin{equation}
  \big[\, \texttt{<image>},\; \underbrace{\texttt{<forensic>} \cdots \texttt{<forensic>}}_{N},\; q \,\big],
\end{equation}
where \texttt{<image>} denotes the visual tokens produced by the MLLM's own vision encoder and $q$ the tokens of the question. At the input-embedding layer, the embeddings of the $N$ placeholders are replaced by $\mathbf{t}_1,\dots,\mathbf{t}_N$. The architecture of the MLLM is otherwise unchanged: the forensic tokens are processed by the same transformer layers as the visual and textual tokens and can be attended to at every layer.

Besides the verdict in the generated text, \ewf{} provides a continuous detection score. Answers start with a fixed opening $o_q$ followed by the verdict (\eg, ``The image looks'' followed by ``real'' or ``fake''). Feeding the input sequence followed by $o_q$, we read the next-token probabilities of the two verdict words and define, with $c = (x, q, o_q)$,
\begin{equation}
  s(x, q) = \frac{P(\text{fake} \mid c)}{P(\text{real} \mid c) + P(\text{fake} \mid c)}\,.
\end{equation}
For a video, $s$ is averaged over its frames.

\subsection{Staged training}
\label{sec:method_training}

Training is performed in four stages; the first three use large-scale detection data and ground the model in deepfake detection, and the last one teaches it to explain. All stages minimize the autoregressive cross-entropy on the answer tokens only, and the detector stays frozen in all of them. The MLLM is adapted with LoRA. 

\textbf{Stage 1: forensic adaptation of the MLLM.} The MLLM alone, without forensic tokens, is first fine-tuned for real/fake classification on FF++~\cite{9010912}, with the question ``Does the image look real/fake?'', followed by the two admissible answers ``The image looks real'' and ``The image looks fake''. This adapts the MLLM to the forensic task and to the answer format using its own visual evidence; the resulting LoRA weights are merged into the model. The detection data match the training regime of the detector: real and manipulated FF++ faces for detectors trained on FF++ forgeries, and real FF++ faces with self-blended images~\cite{9880195} for detectors trained on them. Since this stage does not involve the detector, it is performed once per training regime and shared by all detectors.

\textbf{Stage 2: adapter alignment.} With the detector and the adapted MLLM frozen, only the adapter is trained, on the same task and data. The adapter thus learns to express the detector's evidence in a form that the forensic-aware MLLM can already use, without changing either model.

\textbf{Stage 3: LLM fine-tuning.} The adapter is then frozen and new LoRA weights are trained on the language model of the MLLM (its vision encoder stays frozen in all stages), on the same task and data, so that it learns to exploit the forensic tokens jointly with the image tokens. Training them in separate stages avoids letting them co-adapt to shortcuts before the forensic mapping is established.

\textbf{Stage 4: explanation fine-tuning.} Finally, \ewf{} is fine-tuned on DD-VQA~\cite{10.1007/978-3-031-73223-2_22}. Every (image, question, reference answer) triple is a training sample, covering both image-level and region-level questions. The LoRA weights from stage 3 are fine-tuned further, jointly with the adapter, whose learning rate is ten times smaller, so that the forensic mapping learned on large-scale detection data does not drift on the small, explanation-oriented dataset.

Since the detector is never updated, its representation of the forensic evidence is preserved by construction, whatever the MLLM learns. Whether the MLLM fully exploits this evidence is an empirical question, which we assess in Sec.~\ref{sec:experiments} by comparing \ewf{} with the coupled detector on the same test data, both after the detection-oriented stages and after learning to explain.

\providecommand{\ewf}{EwF}
\providecommand{\sig}[1]{\textsuperscript{#1}}

\section{Experiments}
\label{sec:experiments}

\subsection{Datasets}
\label{sec:exp_data}

All detectors and all detection-oriented training stages of \ewf{} use the FF++ dataset~\cite{9010912}, which contains 1,000 real videos and their manipulated versions produced by five methods (Deepfakes, Face2Face, FaceSwap, NeuralTextures and FaceShifter). We use the official 720/140/140 split into training, validation and test videos, and run experiments at c23 (high quality) and c40 (low quality) compression levels. Training uses all five manipulation methods, while test results are reported on the four original ones (140 real and 560 fake videos), as in the standard FF++ protocol and in the evaluation of LAA-Net~\cite{10657482}. For cross-dataset evaluation, the models trained on FF++ c23 are assessed on the test sets of Celeb-DF (CDF)~\cite{9156368} (178 real and 340 fake videos), DFDC~\cite{DFDC2020} (2,418 real and 2,408 fake videos) and WildDeepfake (WDF)~\cite{10.1145/3394171.3413769} (396 real and 410 fake videos).

For explanation, we use DD-VQA~\cite{10.1007/978-3-031-73223-2_22}, the standard benchmark for explainable face forgery detection~\cite{11092358,kundu-etal-2026-forge,11492581,NEURIPS2025_78a2ed8e}. DD-VQA contains 2,968 face images extracted from FF++ videos and 14,782 question--answer pairs, written by three annotators per image, about the authenticity of the whole face and of individual facial regions (skin, eyebrows, eyes, nose and mouth), each with a textual justification. We use the official split (242 test images, 1,223 test question--answer pairs), with frames at both c23 and c40.

\subsection{Evaluation protocols}
\label{subsec:data_eval_prot}

Deepfake detection is evaluated at the video level using 16 frames sampled uniformly from each test video. The video-level score is the mean of the frame scores, \ie, the detector's fake probability or, for \ewf{}, the score $s(x,q)$ of Sec.~\ref{sec:method_injection}. On these datasets, \ewf{} is evaluated after stage~3, and on DD-VQA after stage~4. We report the video-level AUC and the accuracy obtained by thresholding the video score at 0.5. Since an \ewf{} variant and its detector are evaluated on the same videos, their AUCs are correlated, and comparing point estimates alone ignores both this correlation and the sampling variability. We therefore compare every \ewf{} variant with the detector it is built on, using DeLong's non-parametric test for correlated AUCs~\cite{delong1988comparing,6851192}. As several comparisons are made, we control the family-wise error rate with the Holm--Bonferroni procedure~\cite{holm1979simple}, which is uniformly more powerful than the Bonferroni correction, applied separately to the intra-dataset and the cross-dataset comparisons, and we verify the confidence intervals of the AUC differences with a paired bootstrap~\cite{EfronTibshirani1993} (2,000 resamples of videos).

On DD-VQA, detection performance is evaluated on the image-level question of each test image, as in M2F2-Det, but without revealing the expected verdict in the question. Accuracy and F1 are computed from the verdict in the generated answer, and AUC is calculated from the score $s(x,q)$. For the detectors alone, accuracy and F1 are obtained by thresholding their fake probability at 0.5. Explanations are generated for all test questions (both image- and region-level) with greedy decoding, and compared with all reference answers of the same question using BLEU~\cite{papineni2002bleu}, METEOR~\cite{banerjee2005meteor}, ROUGE-L~\cite{lin2004rouge}, CIDEr~\cite{vedantam2015cider} and SPICE~\cite{anderson2016spice}, computed with the COCO caption evaluation toolkit.

\subsection{Implementation details}
\label{sec:exp_impl}

We use three detectors of different architectures, built with the PyTorch Image Models library (timm)~\cite{rw2019timm} from ImageNet-pretrained weights and trained on FF++ at each compression level: SwinV2-B~\cite{liu2022swinv2} ($N = 64$ forensic tokens), XceptionNet~\cite{chollet2017xception} ($N = 100$) and EfficientNet-B4~\cite{tan2019efficientnet} ($N = 144$). We also use LAA-Net~\cite{10657482} with its released weights ($N = 144$); for its alignment stages, self-blended images are generated from the real FF++ c23 training and validation frames, and no manipulated FF++ video is used for training or model selection.

The MLLM is Qwen3-VL-4B-Instruct~\cite{Qwen3-VL} ($d = 2560$), and the adapter has a hidden size $h = 2560$. LoRA is applied to the attention and MLP projections of the language model (rank 16, $\alpha = 32$, dropout 0.01), while the vision encoder is frozen. Every stage runs for five epochs with AdamW, a learning rate of $10^{-4}$ (adapter: $10^{-5}$ in the explanation stage), 5\% warm-up, cosine decay, an effective batch size of 32 and bfloat16 precision; the detection-oriented training stages use identity-matched 1:1 real/fake frame sampling. The best epoch of every training stage is selected based on validation data only: by video-level AUC on FF++ (on self-blended images for LAA-Net) for the detection-oriented training stages, and by the AUC of $s(x,q)$ on a held-out DD-VQA validation split for the explanation fine-tuning stage. All experiments run on a single NVIDIA RTX 3090 GPU: since the forensic adaptation is shared, integrating a new detector requires 22--27\,hrs of training (stages 2 to 4), and at inference \ewf{} computes the detection score in about 0.15\,sec. and generates a full explanation in about 3.8\,sec. per question. Further computational cost details are provided in the supplementary material. The code will be made publicly available upon acceptance.

\subsection{Experimental results}

\subsubsection{Deepfake detection}
\label{sec:exp_detection}

\begin{table*}[t]
\caption{Intra- and cross-dataset video-level detection performance. \ewf{} variants are evaluated after stage 3. Bold font shows the best result within each detector group. Stars indicate that the difference in terms of AUC between \ewf{} and its coupled detector is statistically significant (DeLong's test, Holm-adjusted within the intra-dataset and the cross-dataset comparisons of the \ewf{} variants aligned in their detector's training regime; $^{**}p<0.01$, $^{***}p<0.001$). Daggers denote the same test for the FF++-aligned \ewf{}\textsubscript{LAA-Net}, adjusted separately ($^{\dagger\dagger\dagger}p<0.001$). LAA-Net is used with its released weights, trained on self-blended images (SBI) from FF++ c0; its \ewf{} variants are aligned either on self-blended images (SBI) or on real and manipulated FF++ videos (FF++). As a single released model is available for LAA-Net, it is evaluated on FF++ at c23 only, the usual setting for cross-dataset evaluation (--: not evaluated).}
\label{tab:detection}
\centering
\small
\begin{tabular}{|l|cc|cc|cccc|}
\hline
 & \multicolumn{4}{c|}{\textbf{Intra-dataset}} & \multicolumn{4}{c|}{\textbf{Cross-dataset (AUC)}} \\ \cline{2-9}
 & \multicolumn{2}{c|}{FF++ c23} & \multicolumn{2}{c|}{FF++ c40} & \multirow{2}{*}{CDF} & \multirow{2}{*}{DFDC} & \multirow{2}{*}{WDF} & \multirow{2}{*}{Avg.} \\
\textbf{Method} & Acc & AUC & Acc & AUC & & & & \\ \hline
SwinV2                             & 0.9629 & 0.9923 & 0.9414 & 0.9717 & 0.8203 & 0.7502 & 0.8040 & 0.7915 \\
\ewf{}\textsubscript{SwinV2}       & \textbf{0.9714} & \textbf{0.9935} & \textbf{0.9486} & \textbf{0.9743} & \textbf{0.8367} & \textbf{0.7621}\sig{***} & \textbf{0.8100} & \textbf{0.8029} \\ \hline
XceptionNet                        & \textbf{0.9814} & \textbf{0.9958} & 0.9471 & 0.9768 & 0.7055 & 0.7376 & \textbf{0.7999} & 0.7477 \\
\ewf{}\textsubscript{XceptionNet}  & 0.9743 & \textbf{0.9958} & \textbf{0.9543} & \textbf{0.9811} & \textbf{0.7151} & \textbf{0.7462}\sig{***} & 0.7984 & \textbf{0.7532} \\ \hline
EfficientNet                       & 0.9700 & 0.9964 & 0.9600 & 0.9785 & 0.7584 & 0.7310 & 0.7648 & 0.7514 \\
\ewf{}\textsubscript{EfficientNet} & \textbf{0.9757} & \textbf{0.9969} & \textbf{0.9629} & \textbf{0.9799} & \textbf{0.7864}\sig{***} & \textbf{0.7465}\sig{***} & \textbf{0.7717} & \textbf{0.7682} \\ \hline
LAA-Net (SBI)                      & 0.5829 & 0.8755 & -- & -- & \textbf{0.9413} & 0.7106 & \textbf{0.8028} & 0.8182 \\
\ewf{}\textsubscript{LAA-Net} (SBI)  & 0.7400 & 0.9072\sig{**} & -- & -- & 0.9202 & 0.7522\sig{***} & 0.7825 & \textbf{0.8183} \\
\ewf{}\textsubscript{LAA-Net} (FF++) & \textbf{0.9086} & \textbf{0.9592}\sig{\dag\dag\dag} & -- & -- & 0.8336\sig{\dag\dag\dag} & \textbf{0.7612}\sig{\dag\dag\dag} & 0.7833 & 0.7927 \\ \hline
\end{tabular}
\end{table*}

Table~\ref{tab:detection} compares every \ewf{} variant with the detector it integrates, intra-dataset on FF++ and cross-dataset on Celeb-DF, DFDC and WildDeepfake. As shown, for the three detectors trained on FF++ forgeries, \ewf{} preserves the near-ceiling intra-dataset performance (all six AUC differences are non-negative and none is significant) and improves cross-dataset generalization; \ie, the AUC increases significantly on DFDC for all three detectors and on Celeb-DF for EfficientNet-B4. The Celeb-DF gains of SwinV2 ($+0.016$) and XceptionNet ($+0.010$) are significant individually ($p<0.01$, with bootstrap 95\% confidence intervals excluding zero), but not after the Holm adjustment over the 12 cross-dataset comparisons ($p_{\text{Holm}}=0.056$ and $0.058$); the remaining differences are not significant. On average over the three unseen datasets, the AUC increases from 0.792 to 0.803 (SwinV2), from 0.748 to 0.753 (XceptionNet) and from 0.751 to 0.768 (EfficientNet-B4). There is no case where an \ewf{} variant, aligned in its detector's training regime, is significantly worse than its detector. \ewf{} also follows the decisions of its coupled detector: for these three detectors, the verdicts coincide on 87.7--99.4\% of the videos (Cohen's $\kappa$ 0.67--0.98), and on the unseen datasets \ewf{} repeats 82--98\% of the detector's errors (see supplementary).

LAA-Net is trained on self-blended images and generalizes better than the supervised detectors (0.941 AUC on Celeb-DF), at the cost of a low in-domain accuracy at the 0.5 threshold (0.583 on FF++). Note that LAA-Net alone is evaluated with its authors' flip test (averaging the predictions for each frame and its horizontal flip), whereas \ewf{}\textsubscript{LAA-Net} gets a single view per frame; this favors LAA-Net in the comparison. When \ewf{}\textsubscript{LAA-Net} is aligned on self-blended images, as the detector itself, it preserves this generalization (no significant difference on Celeb-DF and WildDeepfake; equal average cross-dataset AUC, 0.818), significantly improves DFDC ($+0.042$) and in-domain detection ($+0.032$ AUC, accuracy from 0.583 to 0.740), without seeing any manipulated FF++ video. When aligned on real and manipulated FF++ videos instead, it gains more in-domain ($+0.084$ AUC) but loses much of the detector's generalization capacity ($-0.108$ AUC on Celeb-DF) and largely departs from its decisions (Cohen's $\kappa$ 0.10--0.29). Compared directly, the two alignments trade $0.052$ in-domain AUC for $0.087$ AUC on Celeb-DF (both $p_{\text{Holm}}<0.001$), with no significant difference on DFDC and WildDeepfake. Hence, the detection data used for the alignment should match the training regime of the coupled detector, so that \ewf{} inherits the detector's generalization capacity rather than overriding it.

\begin{table*}[t]
\caption{Detection and explanation performance on DD-VQA (c40). Results marked with $\nabla$ are taken from~\cite{10.1007/978-3-031-73223-2_22} and those marked with $^\blacktriangle$ from~\cite{kundu-etal-2026-forge}. Results marked with $^*$ were computed using the publicly released model and evaluation scripts of the authors of M2F2-Det, on our c40 face images. Non-reported scores are marked with ``-''. Best results are in bold and second-best results are underlined.}
\label{tab:pc_ddvqa_c40}
\centering
\resizebox{\textwidth}{!}{%
\begin{tabular}{|l|c|c|c|c|c|c|c|c|c|c|}
\hline
 & \multicolumn{2}{c|}{\textbf{Backbone Model(s) Details}} & \multicolumn{3}{c|}{\textbf{Detection Performance}} & \multicolumn{5}{c|}{\textbf{Explanation Performance}} \\ \hline
\textbf{Method} & Main Model(s) & \begin{tabular}[c]{@{}c@{}}\# Params\\ Total / Trainable\end{tabular} & Accuracy & F1 & AUC & BLEU-4 & ROUGE-L & CIDEr & METEOR & SPICE \\ \hline
BLIP$^\nabla$~\cite{10.1007/978-3-031-73223-2_22} \footnotesize{\textcolor{gray}{ECCV'24}} & BLIP-Base & 230M / 230M & 0.8168 & 0.8642 & - & 0.3569 & 0.5664 & 1.8177 & 0.3301 & 0.6658 \\
BLIP-T$^\nabla$~\cite{10.1007/978-3-031-73223-2_22} \footnotesize{\textcolor{gray}{ECCV'24}} & BLIP-Base & 230M / 230M & 0.8365 & 0.8758 & - & 0.3714 & 0.5774 & 1.8715 & 0.3349 & 0.6710 \\
BLIP-I$^\nabla$~\cite{10.1007/978-3-031-73223-2_22} \footnotesize{\textcolor{gray}{ECCV'24}} & BLIP-Base & 230M / 230M & 0.8487 & 0.8836 & - & 0.3800 & 0.5882 & 1.8931 & 0.3419 & 0.6788 \\
BLIP-TI$^\nabla$~\cite{10.1007/978-3-031-73223-2_22} \footnotesize{\textcolor{gray}{ECCV'24}} & BLIP-Base & 230M / 230M & 0.8749 & 0.9007 & - & 0.4075 & 0.6085 & 2.0567 & 0.3463 & 0.6915 \\
ForgerySleuth$^\blacktriangle$~\cite{sun2025forgerysleuth} \footnotesize{\textcolor{gray}{NeurIPS'25}} & CLIP ViT-L, SAM ViT-H, Vicuna-7B & 8.1B / 440M & 0.8942 & - & 0.9318 & 0.3812 & 0.5854 & 1.9486 & - & - \\
FakeShield$^\blacktriangle$~\cite{ICLR2025_4d4e0ab9} \footnotesize{\textcolor{gray}{ICLR'25}} & CLIP ViT-L, SAM ViT-H, Vicuna-13B & 14B / 400M & {\ul 0.9085} & - & 0.9391 & 0.3947 & 0.5982 & 2.0143 & - & - \\
FORGE$^\blacktriangle$~\cite{kundu-etal-2026-forge} \footnotesize{\textcolor{gray}{EMNLP'26}} & SigLIP-So400m/14, DINOv2, Gemma2 & 3.3B / 2.6B & \textbf{0.9412} & - & \textbf{0.9539} & 0.4304 & 0.6285 & 2.6321 & - & - \\
M2F2-Det$^*$~\cite{11092358} \footnotesize{\textcolor{gray}{CVPR'25}} & CLIP, EfficientNet-B4, Vicuna-7B & 7.5B / 85M & 0.8951 & 0.8905 & - & 0.4656 & 0.6425 & 3.3546 & - & 0.6343 \\ \hline
\ewf{}\textsubscript{SwinV2} & SwinV2-Base, Qwen3-VL-4B & 4.57B / 42M & 0.8554 & 0.8961 & 0.9398 & \textbf{0.6092} & 0.7424 & 3.8230 & 0.4301 & {\ul 0.6953} \\
\ewf{}\textsubscript{XceptionNet} & XceptionNet, Qwen3-VL-4B & 4.50B / 45M & 0.8678 & {\ul 0.9048} & {\ul 0.9519} & {\ul 0.5957} & {\ul 0.7526} & {\ul 3.8713} & {\ul 0.4391} & 0.6924 \\
\ewf{}\textsubscript{EfficientNet} & EfficientNet-B4, Qwen3-VL-4B & 4.50B / 44M & 0.9008 & \textbf{0.9286} & 0.9464 & 0.5940 & \textbf{0.7682} & \textbf{3.9257} & \textbf{0.4601} & \textbf{0.7051} \\ \hline
\end{tabular}}
\end{table*}

\begin{table*}[t]
\caption{Detection and explanation performance on DD-VQA (c23). Results marked with $^\ddagger$ are taken from~\cite{11492581} and those marked with $^\blacktriangle$ from~\cite{kundu-etal-2026-forge}. Non-reported scores are marked with ``-''. Best results are in bold and second-best results are underlined.}
\label{tab:pc_ddvqa_c23}
\centering
\resizebox{\textwidth}{!}{%
\begin{tabular}{|l|c|c|c|c|c|c|c|c|c|c|}
\hline
 & \multicolumn{2}{c|}{\textbf{Backbone Model(s) Details}} & \multicolumn{3}{c|}{\textbf{Detection Performance}} & \multicolumn{5}{c|}{\textbf{Explanation Performance}} \\ \hline
\textbf{Method} & Main Model(s) & \begin{tabular}[c]{@{}c@{}}\# Params\\ Total / Trainable\end{tabular} & Accuracy & F1 & AUC & BLEU-4 & ROUGE-L & CIDEr & METEOR & SPICE \\ \hline
AuthGuard$^\ddagger$~\cite{11492581} \footnotesize{\textcolor{gray}{WACV'26}} & CLIP ViT-L, RoBERTa, Vicuna-7B & 7.5B / 300M & 0.9084 & - & - & 0.4980 & 0.6950 & 3.3050 & 0.4010 & - \\
FORGE$^\blacktriangle$~\cite{kundu-etal-2026-forge} \footnotesize{\textcolor{gray}{EMNLP'26}} & SigLIP-So400m/14, DINOv2, Gemma2 & 3.3B / 2.6B & 0.9361 & - & - & 0.4269 & 0.6103 & 2.6057 & - & - \\ \hline
\ewf{}\textsubscript{SwinV2} & SwinV2-Base, Qwen3-VL-4B & 4.57B / 42M & 0.9463 & 0.9600 & {\ul 0.9868} & \textbf{0.6796} & \textbf{0.8027} & \textbf{4.5941} & 0.4604 & \textbf{0.7343} \\
\ewf{}\textsubscript{XceptionNet} & XceptionNet, Qwen3-VL-4B & 4.50B / 45M & \textbf{0.9711} & \textbf{0.9781} & 0.9852 & {\ul 0.6457} & {\ul 0.8022} & {\ul 4.4477} & {\ul 0.4660} & {\ul 0.7259} \\
\ewf{}\textsubscript{EfficientNet} & EfficientNet-B4, Qwen3-VL-4B & 4.50B / 44M & {\ul 0.9669} & {\ul 0.9750} & \textbf{0.9870} & 0.6202 & 0.7995 & 4.2258 & \textbf{0.4784} & 0.7140 \\ \hline
\end{tabular}}
\end{table*}

\begin{table*}[t]
\caption{Effect of training data on DD-VQA (c40). Bold font shows the best result within each method. For the detectors, accuracy and F1 are computed at a 0.5 threshold on their fake probability. The AUC difference between each detector and its \ewf{} variant trained on FF++ \& DD-VQA is not significant (DeLong's test on the 242 test images, Holm-adjusted over the six comparisons at both compression levels).}
\label{tab:training_ddvqa_c40}
\centering
\resizebox{\textwidth}{!}{%
\footnotesize
\begin{tabular}{|l|c|ccc|ccccc|}
\hline
 & & \multicolumn{3}{c|}{\textbf{Detection Performance}} & \multicolumn{5}{c|}{\textbf{Explanation Performance}} \\ \hline
\textbf{Method} & \textbf{Training data} & Accuracy & F1 & AUC & BLEU-4 & ROUGE-L & CIDEr & METEOR & SPICE \\ \hline
SwinV2       & FF++ & 0.8678 & 0.9006 & 0.9341 & n/a & n/a & n/a & n/a & n/a \\
XceptionNet  & FF++ & 0.8926 & 0.9187 & 0.9486 & n/a & n/a & n/a & n/a & n/a \\
EfficientNet & FF++ & 0.9132 & 0.9354 & 0.9587 & n/a & n/a & n/a & n/a & n/a \\ \hline
\multirow{4}{*}{Qwen3-VL-4B}
  & --             & 0.6322 & 0.7003 & 0.6406 & 0.0372 & 0.1768 & 0.0000 & 0.1663 & 0.1330 \\
  & FF++            & \textbf{0.8099} & \textbf{0.8631} & \textbf{0.8871} & 0.0978 & 0.2608 & 0.1135 & 0.1656 & 0.2547 \\
  & DD-VQA          & 0.7479 & 0.8211 & 0.8140 & \textbf{0.5438} & \textbf{0.7179} & \textbf{3.5603} & \textbf{0.4278} & \textbf{0.6875} \\
  & FF++ \& DD-VQA  & 0.7107 & 0.8098 & 0.8631 & 0.5159 & 0.6805 & 3.0883 & 0.4089 & 0.6575 \\ \hline
\multirow{3}{*}{\ewf{}\textsubscript{SwinV2}}
  & FF++            & \textbf{0.8595} & \textbf{0.8970} & \textbf{0.9439} & 0.1136 & 0.2764 & 0.1738 & 0.1538 & 0.3226 \\
  & DD-VQA          & 0.7893 & 0.8539 & 0.8497 & 0.5485 & 0.7209 & 3.5168 & \textbf{0.4351} & 0.6871 \\
  & FF++ \& DD-VQA  & 0.8554 & 0.8961 & 0.9398 & \textbf{0.6092} & \textbf{0.7424} & \textbf{3.8230} & 0.4301 & \textbf{0.6953} \\ \hline
\multirow{3}{*}{\ewf{}\textsubscript{XceptionNet}}
  & FF++            & \textbf{0.8843} & \textbf{0.9157} & 0.9434 & 0.0749 & 0.2377 & 0.3208 & 0.0973 & 0.2694 \\
  & DD-VQA          & 0.8347 & 0.8802 & 0.8893 & 0.5612 & 0.7383 & 3.6345 & 0.4389 & 0.6835 \\
  & FF++ \& DD-VQA  & 0.8678 & 0.9048 & \textbf{0.9519} & \textbf{0.5957} & \textbf{0.7526} & \textbf{3.8713} & \textbf{0.4391} & \textbf{0.6924} \\ \hline
\multirow{3}{*}{\ewf{}\textsubscript{EfficientNet}}
  & FF++            & \textbf{0.9091} & \textbf{0.9317} & \textbf{0.9476} & 0.1247 & 0.2979 & 0.0959 & 0.1886 & 0.2621 \\
  & DD-VQA          & 0.8306 & 0.8791 & 0.8778 & 0.5608 & 0.7409 & 3.6323 & 0.4435 & 0.6870 \\
  & FF++ \& DD-VQA  & 0.9008 & 0.9286 & 0.9464 & \textbf{0.5940} & \textbf{0.7682} & \textbf{3.9257} & \textbf{0.4601} & \textbf{0.7051} \\ \hline
\end{tabular}}
\end{table*}

\subsubsection{Explainable deepfake detection}
\label{sec:exp_sota}

Having shown that the detection-oriented stages preserve the detection capability of the coupled detectors, we now assess the full framework on DD-VQA. Tables~\ref{tab:pc_ddvqa_c40} and~\ref{tab:pc_ddvqa_c23} compare \ewf{} with state-of-the-art explainable deepfake detection methods \cite{10.1007/978-3-031-73223-2_22,sun2025forgerysleuth,ICLR2025_4d4e0ab9,kundu-etal-2026-forge,11092358,11492581} on DD-VQA. In explanation, \ewf{} clearly outperforms all of them at both compression levels
(Fig.~\ref{fig:qual_main}; more examples in the supplementary material); every explanation metric is best for one of its variants and second best for another. At c40, the best \ewf{} variant improves on M2F2-Det by $+0.14$ BLEU-4, $+0.13$ ROUGE-L and $+0.57$ CIDEr, although M2F2-Det was evaluated with its own protocol, in which the questions reveal the expected verdict\footnote{See the M2F2-Det repository: \url{https://github.com/CHELSEA234/M2F2_Det/tree/main}}; at c23, it improves on the strongest competitor, AuthGuard, by $+0.18$ BLEU-4 ($0.680$ vs.\ $0.498$), $+0.11$ ROUGE-L and $+1.29$ CIDEr ($4.59$ vs.\ $3.31$). These gains are obtained with a smaller model (4.5--4.6B parameters, of which only 42--45M are trained) than most competitors, \eg, 7.5B/85M for M2F2-Det and 3.3B/2.6B for FORGE. In terms of detection performance, \ewf{} achieves the best F1 and the second-best AUC at c40, with FORGE remaining ahead in accuracy and AUC, and the best accuracy at c23 (\eg, 0.971 for \ewf{}\textsubscript{XceptionNet} vs.\ 0.936 for FORGE).

\begin{figure}[t]
\centering
\scriptsize
\begin{tabular}{@{}m{0.22\linewidth}@{\hspace{0.02\linewidth}}m{0.76\linewidth}@{}}
\includegraphics[width=\linewidth]{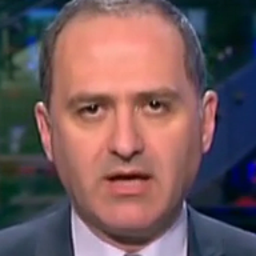} &
\textbf{Q:} Does the person's nose look real/fake? \hfill (label: \textbf{fake}) \newline
\textbf{Ref.:} \emph{The person's nose looks fake. The person's nose looks very fake because
the nose lacks of subtle details, such as pores, fine lines.} \newline
\textbf{\ewf{}\textsubscript{SwinV2}} \textcolor{green!50!black}{$\checkmark$}\textbf{:} The
person's nose looks fake. The person's nose looks a bit fake because the nose lookes curved. \newline
\textbf{Qwen3-VL-4B} \textcolor{red!70!black}{$\times$}\textbf{:} The person's nose looks real.
The person has naturally straight nose.
\end{tabular}
\caption{Answers of \ewf{}\textsubscript{SwinV2} and Qwen3-VL-4B (trained on DD-VQA) to a
  region-level question on DD-VQA (c23), reproduced verbatim. $\checkmark$/$\times$:
  correct/wrong verdict. More examples, including failure cases, are given in the
  supplementary material.}
\label{fig:qual_main}
\end{figure}

\subsubsection{Ablation study on training data}
\label{sec:exp_ablation}

\begin{table*}[t]
\caption{Effect of training data on DD-VQA (c23). Bold font shows the best result within each method. For the detectors, accuracy and F1 are computed at a 0.5 threshold on their fake probability. The AUC difference between each detector and its \ewf{} variant trained on FF++ \& DD-VQA is not significant (DeLong's test on the 242 test images, Holm-adjusted over the six comparisons at both compression levels).}
\label{tab:training_ddvqa_c23}
\centering
\resizebox{\textwidth}{!}{%
\footnotesize
\begin{tabular}{|l|c|ccc|ccccc|}
\hline
 & & \multicolumn{3}{c|}{\textbf{Detection Performance}} & \multicolumn{5}{c|}{\textbf{Explanation Performance}} \\ \hline
\textbf{Method} & \textbf{Training data} & Accuracy & F1 & AUC & BLEU-4 & ROUGE-L & CIDEr & METEOR & SPICE \\ \hline
SwinV2       & FF++ & 0.9545 & 0.9649 & 0.9833 & n/a & n/a & n/a & n/a & n/a \\
XceptionNet  & FF++ & 0.9752 & 0.9810 & 0.9890 & n/a & n/a & n/a & n/a & n/a \\
EfficientNet & FF++ & 0.9504 & 0.9615 & 0.9914 & n/a & n/a & n/a & n/a & n/a \\ \hline
\multirow{4}{*}{Qwen3-VL-4B}
  & --             & 0.4587 & 0.3417 & 0.6236 & 0.0406 & 0.1826 & 0.0005 & 0.1664 & 0.1355 \\
  & FF++            & \textbf{0.8471} & \textbf{0.8862} & \textbf{0.9272} & 0.0783 & 0.2491 & 0.0919 & 0.1726 & 0.2226 \\
  & DD-VQA          & 0.7479 & 0.8100 & 0.8304 & \textbf{0.5730} & 0.7153 & \textbf{3.6112} & 0.4063 & 0.6817 \\
  & FF++ \& DD-VQA  & 0.8099 & 0.8671 & 0.9164 & 0.5425 & \textbf{0.7301} & 3.5766 & \textbf{0.4429} & \textbf{0.6839} \\ \hline
\multirow{3}{*}{\ewf{}\textsubscript{SwinV2}}
  & FF++            & \textbf{0.9587} & \textbf{0.9684} & \textbf{0.9898} & 0.1476 & 0.3044 & 0.2946 & 0.1533 & 0.3943 \\
  & DD-VQA          & 0.9091 & 0.9333 & 0.9609 & 0.6179 & 0.7736 & 4.0572 & 0.4448 & 0.7126 \\
  & FF++ \& DD-VQA  & 0.9463 & 0.9600 & 0.9868 & \textbf{0.6796} & \textbf{0.8027} & \textbf{4.5941} & \textbf{0.4604} & \textbf{0.7343} \\ \hline
\multirow{3}{*}{\ewf{}\textsubscript{XceptionNet}}
  & FF++            & \textbf{0.9752} & \textbf{0.9811} & \textbf{0.9857} & 0.0451 & 0.1752 & 0.0166 & 0.1418 & 0.3811 \\
  & DD-VQA          & 0.9256 & 0.9458 & 0.9768 & 0.6147 & 0.7850 & 4.1475 & \textbf{0.4678} & 0.7175 \\
  & FF++ \& DD-VQA  & 0.9711 & 0.9781 & 0.9852 & \textbf{0.6457} & \textbf{0.8022} & \textbf{4.4477} & 0.4660 & \textbf{0.7259} \\ \hline
\multirow{3}{*}{\ewf{}\textsubscript{EfficientNet}}
  & FF++            & 0.9628 & 0.9709 & 0.9857 & 0.0989 & 0.3556 & 0.9222 & 0.1809 & 0.3631 \\
  & DD-VQA          & 0.8843 & 0.9181 & 0.9661 & 0.5816 & 0.7570 & 3.8666 & 0.4511 & 0.6964 \\
  & FF++ \& DD-VQA  & \textbf{0.9669} & \textbf{0.9750} & \textbf{0.9870} & \textbf{0.6202} & \textbf{0.7995} & \textbf{4.2258} & \textbf{0.4784} & \textbf{0.7140} \\ \hline
\end{tabular}}
\end{table*}

We assess the proposed training scheme on DD-VQA, by training Qwen3-VL-4B and the \ewf{} variants on detection data only (FF++), explanation data only (DD-VQA), or both (FF++ \& DD-VQA). The results in Tables~\ref{tab:training_ddvqa_c40} and~\ref{tab:training_ddvqa_c23} show that, without fine-tuning, Qwen3-VL-4B is close to chance in detection (0.64/0.62 AUC at c40/c23) and its free-form answers barely match the references (BLEU-4 $\approx 0.04$). Models trained on FF++ only detect well but answer with a bare verdict, which yields low explanation scores. Models trained on DD-VQA only explain well but detect worse; for \ewf{}, the AUC drops to 0.85--0.89 at c40 and to 0.96--0.98 at c23, compared with 0.94--0.95 and 0.99, respectively, when FF++ is also used. Hence, \textbf{each capability, detection and explanation, requires its own training data}.

Moreover, for every \ewf{} variant and at both compression levels, the full scheme gives the best explanation results on all metrics except METEOR for two variants (where the difference is at most 0.005), while its detection performance remains close to that of the FF++-only model. In addition, it does not differ significantly from that of the detector alone; on the 242 test images, the AUC differences range from $-0.012$ to $+0.006$ (DeLong's test, all $p_{\text{Holm}}=1$), and the verdict in the generated answer agrees with the detector on 88.0--98.8\% of the test images ($\kappa$ 0.71--0.97). So, \textbf{learning to explain preserves the detection capability of the coupled detector}.

For Qwen3-VL-4B alone, fine-tuning on FF++ before DD-VQA lowers explanation quality and accuracy at c40, despite a higher AUC, and gives mixed results at c23, whereas for every \ewf{} variant it improves explanation at both compression levels on almost all metrics. Hence, \textbf{training on detection data (FF++) benefits explanation only when the MLLM receives the forensic tokens.}

Compared with Qwen3-VL-4B trained on the same data (FF++ \& DD-VQA), every \ewf{} variant is significantly better at both compression levels, in detection ($+0.07$ to $+0.09$ AUC; DeLong's test) and in explanation ($+0.65$ to $+1.02$ CIDEr; image-level paired permutation test~\cite{riezler2005some}), with $p_{\text{Holm}}<0.001$ in all 12 comparisons (see supplementary). The advantage also holds with DD-VQA data only (AUC from 0.81 to 0.85--0.89 at c40 and from 0.83 to 0.96--0.98 at c23) and at the region level: on every facial-region question, the \ewf{} variants trained with the full scheme give clearly more accurate verdicts than both Qwen3-VL-4B models (0.79--0.90 vs.\ at most 0.78 at c40, and 0.93--0.97 vs.\ at most 0.81 at c23) and, for the best variant, a higher CIDEr (see supplementary). These findings indicate that \textbf{the forensic tokens of the frozen detector significantly improve both detection and explanation over the same MLLM trained on the same data.}
\section{Conclusions}
\label{sec:conclusion}

We presented \ewf{}, a framework that couples a frozen, off-the-shelf deepfake detector with an MLLM through per-location forensic tokens, so that the detector's spatial evidence reaches the language model and any detector exposing a feature map can be plugged in, by training only a lightweight adapter and performing low-rank adaptation on the MLLM. Combined with a staged training scheme that grounds the framework in detection before teaching it to explain, \ewf{} preserves or significantly improves the detection performance of every coupled detector, when aligned on data that match its training regime, and produces explanations that outperform those of state-of-the-art methods on the DD-VQA dataset and are significantly better than those of the same MLLM trained on the same data without forensic tokens. Our experiments also show that the detection data used for alignment should match the detector's training regime, so that \ewf{} inherits the detector's generalization rather than overriding it. Our study has limitations: we use a single MLLM and a single explanation benchmark, assess explanations with reference-based text metrics rather than with human judgments, and do not measure how faithfully they reflect the detector's evidence, beyond the consistency of their verdicts with the detector (see supplementary). Future work includes quantifying the contribution of the spatial layout of the forensic tokens to region-level explanations, and extending \ewf{} to support the explainable detection of fully synthetic content.

\section*{Acknowledgments}
This work was supported by the European Union (EU) Horizon Europe Program under Grant 101214398 ELLIOT.

\clearpage
\setcounter{page}{1}
\setcounter{dbltopnumber}{3}
\renewcommand{\dbltopfraction}{0.9}

\maketitlesupplementary

\providecommand{\ewf}{EwF}

\noindent The numbering of sections, tables and figures continues from the main paper, and the reference numbers refer to its bibliography.

\section{Additional implementation details}
\label{sec:supp_impl}

\paragraph{Face crops.}
For all the videos of the employed datasets, and for the DD-VQA images, we used face crops, one per frame: faces are detected with RetinaFace~\cite{9157330}, the detected box is enlarged by 30\%, and the crop is stored at its native resolution, which depends on the size of the face; each model then resizes it with its own
preprocessing. For FF++, Celeb-DF, DFDC and WildDeepfake, 16 frames per video are scored, chosen evenly over the frames of the video ordered by time (frame $\lfloor (j+0.5)\,n/16 \rfloor$, $j=0,\dots,15$, for a video with $n$ available frames; all frames when $n \le 16$). In DFDC, frames are grouped into videos by their video ID.

\paragraph{Detector preprocessing.}
Each detector receives the face crop preprocessed exactly as in its own training and evaluation, using the timm (PyTorch Image Models) data configuration stored with its checkpoint: the crop is resized with bicubic interpolation so that its shorter side equals the input size divided by the crop ratio, center-cropped to the input size, and normalized (Tab.~\ref{tab:supp_preproc}). The MLLM receives the same face crop through its own image processor, with its default settings, so the two branches see the same image at their native resolutions.

\begin{table*}[t]
\centering
\caption{Detector input preprocessing (timm data configurations).}
\label{tab:supp_preproc}
\begin{tabular}{|l|c|c|c|c|}
\hline
\textbf{Detector} & \textbf{Input} & \textbf{Crop ratio} & \textbf{Mean} & \textbf{Std} \\ \hline
SwinV2-B (\texttt{swinv2\_base\_window8\_256}) & $256^2$ & 0.900 & ImageNet & ImageNet \\
XceptionNet (\texttt{legacy\_xception}) & $299^2$ & 0.8975 & (0.5, 0.5, 0.5) & (0.5, 0.5, 0.5) \\
EfficientNet-B4 (\texttt{tf\_efficientnet\_b4.ns\_jft\_in1k}) & $380^2$ & 0.922 & ImageNet & ImageNet \\ \hline
\end{tabular}
\end{table*}

For LAA-Net, we follow its authors' test pipeline: the face crop is resized to $256 \times 256$, cropped by a margin of 5 pixels on each side, warped to $384 \times 384$ with an affine transform (bilinear interpolation), scaled to $[0,1]$ and normalized with mean and standard deviation 0.5. Inside \ewf{}, LAA-Net runs in float32 with mixed precision disabled; the fused E-FPN map, read at the input of its classification head, is average-pooled to $12 \times 12$. When LAA-Net is evaluated alone, we use its authors' flip test (the classification logits of each frame and its horizontal flip are averaged) and $p_{\text{fake}} = \sigma(\text{logit})$; \ewf{} sees a single view per frame.

\paragraph{Detector training.}
SwinV2-B, XceptionNet and EfficientNet-B4 are fine-tuned from their ImageNet-pretrained timm weights as binary (real/fake) classifiers on the FF++ training videos of each compression level (with the same settings at c23 and c40), with all five manipulation methods and up to 16 frames per video, for 10 epochs, with a batch size of 16 and a learning rate of $3 \times 10^{-5}$ for SwinV2-B, and a batch size of 32 and a learning rate of $2 \times 10^{-4}$ for XceptionNet and EfficientNet-B4 (seed 42). Training uses AdamW (weight decay 0.01, not applied to normalization and bias parameters), a linear warm-up over 5\% of the steps followed by cosine decay to 1\% of the learning rate, gradient clipping at norm 1.0, the cross-entropy loss and bfloat16 mixed precision. Frames are sampled with the same identity-matched 1:1 real/fake scheme as for \ewf{}, redrawn every epoch. Augmentations are kept light, so that the face stays in frame and keeps its scale: the crop is resized as at evaluation, then cropped at a random position instead of the center (a shift of at most about 10\% of the input size), followed by random horizontal flipping, color jitter (brightness, contrast and saturation 0.1, hue 0.02; probability 0.5) and JPEG re-compression at a random quality between 60 and 95 (probability 0.3). Random resized crops are not used, since they can cut out or rescale the blending boundaries that deepfake detectors rely on. The epoch is selected by video-level AUC on the FF++ validation set (16 frames per video), as for \ewf{}.

\paragraph{Training of \ewf{}.}
All training stages use an effective batch size of 32 (4 samples per step, 8 gradient accumulation steps). In the detection-oriented training stages, each epoch draws up to 16 random frames from every real training video and the same number of frames from its manipulated versions, split evenly across the manipulation methods; this identity-matched 1:1 sampling is redrawn every epoch. The DD-VQA validation split used for model selection is held out from the DD-VQA training set and has the same size as the test split. Generated answers are limited to 256 new tokens (greedy decoding).

\paragraph{Self-blended images.}
For the alignment of \ewf{}\textsubscript{LAA-Net} on self-blended images, every real FF++ c23 training and validation frame yields two blended variants, generated with LAA-Net's implementation of SBI~\cite{9880195}: 81 facial landmarks (dlib), a randomly deformed convex hull as blending mask, and dynamic blending of the frame with a transformed copy of itself. Frames in which no face is detected (about 1\%) are skipped. Real and blended images are sampled 1:1 within each identity, as for FF++.

\paragraph{Detection score and verdict.}
The detection score $s(x,q)$ is read at the first token where the answers ``\ldots real'' and ``\ldots fake'' differ after tokenization, in a single forward pass with the answer opening appended to the prompt. For DD-VQA, the opening of each question type (\eg, ``The image looks'', ``The person's nose looks'') is taken as the most frequent prefix before the verdict in the reference answers of the training split. The verdict of a generated answer is the first ``real'' or ``fake'' in its first sentence; an answer without a verdict is counted as wrong.

\paragraph{Text metrics.}
BLEU-4, METEOR, ROUGE-L, CIDEr and SPICE are computed with the COCO caption evaluation toolkit (\texttt{pycocoevalcap}) after PTB (Penn Treebank style) tokenization, each generated answer against all reference answers of its question; empty answers are replaced by the word ``empty''. M2F2-Det was evaluated with the officially released model and evaluation scripts on our c40 face images.

\paragraph{Statistical tests.}
DeLong's test uses the fast algorithm of~\cite{6851192}; the paired bootstrap resamples videos (or DD-VQA images) with replacement, 2,000 times, with both models evaluated on the same resample. Holm's procedure is applied within each family fixed in advance: (i)~intra-dataset, \ewf{} vs.\ its detector on FF++ (3 detectors $\times$ 2 compression levels, plus the SBI-aligned \ewf{}\textsubscript{LAA-Net} at c23; 7 tests); (ii)~cross-dataset, the same comparisons on Celeb-DF, DFDC and WildDeepfake (12 tests); (iii)~the FF++-aligned \ewf{}\textsubscript{LAA-Net} vs.\ LAA-Net (4 tests); (iv)~FF++-aligned vs.\ SBI-aligned \ewf{}\textsubscript{LAA-Net} (4 tests); (v)~on DD-VQA, each detector vs.\ its \ewf{} variant trained on FF++ \& DD-VQA (3 detectors $\times$ 2 compression levels; 6 tests); and (vi)~on DD-VQA, each \ewf{} variant trained on FF++ \& DD-VQA vs.\ Qwen3-VL-4B trained on the same data, in detection (AUC, DeLong's test; 6 tests) and in explanation (CIDEr; 6 tests). For the explanation tests, CIDEr is computed per test question (image- and region-level), each answer against all reference answers of its question; these scores average to the reported CIDEr. Since the questions about one image are not independent, the image is the unit of resampling: the paired permutation test~\cite{riezler2005some} swaps the two models' answers to all questions of an image with probability 1/2 (10,000 permutations), and the 95\% confidence intervals come from 10,000 image-level bootstrap resamples. In (vi), \ewf{} is significantly better in all cases ($p_{\text{Holm}}<0.001$), by $+0.07$ to $+0.09$ AUC and $+0.65$ to $+1.02$ CIDEr. Two further comparisons were tested in the same way, each as a separate family. Against Qwen3-VL-4B trained on DD-VQA only, \ewf{} is significantly better in AUC in all cases ($+0.13$ to $+0.16$, $p_{\text{Holm}}<0.001$) and in CIDEr at c23 ($+0.61$ to $+0.98$, $p_{\text{Holm}}<0.01$), while its CIDEr gain at c40 ($+0.26$ to $+0.37$) is not significant ($p_{\text{Holm}}=0.17$). Against the same \ewf{} variant trained on DD-VQA only, the full scheme gains $+0.24$ to $+0.54$ CIDEr in all six cases, which is significant after the Holm adjustment for \ewf{}\textsubscript{SwinV2} at c23 ($p_{\text{Holm}}<0.001$) and individually ($p<0.05$) for four of the remaining five.

\paragraph{Parameter counts.}
The total number of parameters of \ewf{} is the sum of Qwen3-VL-4B (4.44B), the detector (SwinV2-B 86.9M, XceptionNet 20.8M, EfficientNet-B4 17.6M), the adapter (9.2M, 11.8M and 11.2M, respectively, depending on the channel dimension $C$) and the LoRA weights (33.0M); the trainable parameters during the explanation fine-tuning stage are those of the adapter and the LoRA weights (42--45M).

\paragraph{Compute.}
All training stages run on a single NVIDIA RTX 3090 GPU (24\,GB). Measured over five epochs, including the validation after every epoch, the forensic adaptation of the MLLM (stage 1) takes 11.4--15.0\,hrs; it does not involve the detector, so it is run once per training regime and compression level and shared by all detectors. The adapter alignment (stage 2) then takes 6.7--8.0\,hrs and the LLM fine-tuning (stage 3) 9.0--10.6\,hrs per detector, increasing slightly with the number of forensic tokens (64 for SwinV2-B, 100 for XceptionNet, 144 for EfficientNet-B4 and LAA-Net) and identical at c23 and c40. Explanation fine-tuning on DD-VQA (stage 4) takes 6.7--7.9\,hrs. Adding a new detector to \ewf{} thus requires 22--27\,hrs of training on a single GPU, as the forensic adaptation is reused. At inference, with a batch size of 1 and including data loading and preprocessing, \ewf{}\textsubscript{EfficientNet} (the variant with the most forensic tokens) computes the detection score $s(x,q)$ in about 0.15\,sec. and generates a full answer (about 40 words on average) in about 3.8\,sec. per DD-VQA question.

\begin{table*}[t]
\caption{Region-level results on DD-VQA (c23), per question type (number of test questions in parentheses): accuracy of the real/fake verdict in the generated answer, with respect to the image label, and CIDEr of the answer. Qwen3-VL-4B and the \ewf{} variants are fine-tuned on the same data; FF++ \& DD-VQA denotes the full training scheme. Bold: best result per column.}
\label{tab:region_ddvqa_c23}
\centering
\resizebox{\textwidth}{!}{%
\begin{tabular}{|l|cc|cc|cc|cc|cc|cc|}
\hline
 & \multicolumn{2}{c|}{\textbf{Image} (242)} & \multicolumn{2}{c|}{\textbf{Skin} (209)} & \multicolumn{2}{c|}{\textbf{Eyebrows} (192)} & \multicolumn{2}{c|}{\textbf{Eyes} (192)} & \multicolumn{2}{c|}{\textbf{Nose} (200)} & \multicolumn{2}{c|}{\textbf{Mouth} (188)} \\
\textbf{Method} & Acc & CIDEr & Acc & CIDEr & Acc & CIDEr & Acc & CIDEr & Acc & CIDEr & Acc & CIDEr \\ \hline
Qwen3-VL-4B (DD-VQA) & 0.7479 & 0.8818 & 0.7321 & 3.4662 & 0.7448 & 3.4024 & 0.7604 & 3.2156 & 0.7500 & 3.5387 & 0.7340 & 3.6017 \\
Qwen3-VL-4B (FF++ \& DD-VQA) & 0.8099 & 0.6125 & 0.8086 & 3.8617 & 0.7812 & 3.4455 & 0.7969 & 3.3183 & 0.8050 & 3.5013 & 0.7872 & 3.0396 \\ \hline
\ewf{}\textsubscript{SwinV2} & 0.9463 & \textbf{1.2363} & 0.9330 & 4.7365 & 0.9375 & \textbf{4.4660} & 0.9375 & 4.0828 & 0.9400 & \textbf{4.6104} & 0.9255 & \textbf{4.7092} \\
\ewf{}\textsubscript{XceptionNet} & \textbf{0.9711} & 1.1538 & \textbf{0.9665} & 4.9564 & \textbf{0.9635} & 3.8283 & \textbf{0.9583} & 3.8984 & \textbf{0.9650} & 4.5568 & \textbf{0.9628} & 4.6374 \\
\ewf{}\textsubscript{EfficientNet} & 0.9669 & 1.0845 & \textbf{0.9665} & \textbf{4.9861} & 0.9479 & 3.4071 & \textbf{0.9583} & \textbf{4.1898} & 0.9600 & 4.6057 & 0.9574 & 3.9789 \\ \hline
\end{tabular}}
\end{table*}

\begin{table*}[t]
\caption{Region-level results on DD-VQA (c40), per question type (number of test questions in parentheses): accuracy of the real/fake verdict in the generated answer, with respect to the image label, and CIDEr of the answer. Qwen3-VL-4B and the \ewf{} variants are fine-tuned on the same data; FF++ \& DD-VQA denotes the full training scheme. Bold: best result per column.}
\label{tab:region_ddvqa_c40}
\centering
\resizebox{\textwidth}{!}{%
\begin{tabular}{|l|cc|cc|cc|cc|cc|cc|}
\hline
 & \multicolumn{2}{c|}{\textbf{Image} (242)} & \multicolumn{2}{c|}{\textbf{Skin} (209)} & \multicolumn{2}{c|}{\textbf{Eyebrows} (192)} & \multicolumn{2}{c|}{\textbf{Eyes} (192)} & \multicolumn{2}{c|}{\textbf{Nose} (200)} & \multicolumn{2}{c|}{\textbf{Mouth} (188)} \\
\textbf{Method} & Acc & CIDEr & Acc & CIDEr & Acc & CIDEr & Acc & CIDEr & Acc & CIDEr & Acc & CIDEr \\ \hline
Qwen3-VL-4B (DD-VQA) & 0.7479 & 0.7154 & 0.7847 & 3.7865 & 0.7344 & 3.1842 & 0.7500 & 3.2351 & 0.7550 & 3.3403 & 0.7234 & 3.4937 \\
Qwen3-VL-4B (FF++ \& DD-VQA) & 0.7107 & 0.6661 & 0.6890 & 3.0711 & 0.6615 & 2.6992 & 0.6875 & 2.5546 & 0.6700 & 2.7790 & 0.6755 & 3.0632 \\ \hline
\ewf{}\textsubscript{SwinV2} & 0.8554 & \textbf{1.0122} & 0.8230 & 3.9938 & 0.7917 & 3.4325 & 0.8125 & 3.1591 & 0.8150 & 3.9680 & 0.8032 & \textbf{3.7898} \\
\ewf{}\textsubscript{XceptionNet} & 0.8678 & 0.7806 & 0.8565 & 4.1654 & 0.8438 & 3.7383 & 0.8594 & \textbf{3.6270} & 0.8600 & 3.8885 & 0.8457 & 3.3180 \\
\ewf{}\textsubscript{EfficientNet} & \textbf{0.9008} & 0.8659 & \textbf{0.8900} & \textbf{4.3999} & \textbf{0.8750} & \textbf{3.7480} & \textbf{0.8958} & 3.5920 & \textbf{0.8850} & \textbf{4.0532} & \textbf{0.8830} & 3.2005 \\ \hline
\end{tabular}}
\end{table*}

\begin{table*}[t]
\caption{Decision consistency between each \ewf{} variant and the detector it integrates: share of test samples on which the two give the same real/fake verdict and, in parentheses, Cohen's $\kappa$. Detection datasets: video-level verdicts (score $\geq 0.5$), models of Table~\ref{tab:detection}. DD-VQA: the verdict in the generated answer to the image-level question vs.\ the detector's verdict on the image, \ewf{} trained on FF++ \& DD-VQA. --: not evaluated.}
\label{tab:supp_consistency}
\centering
\resizebox{\textwidth}{!}{%
\footnotesize
\begin{tabular}{|l|cc|ccc|cc|}
\hline
 & \multicolumn{2}{c|}{\textbf{Intra-dataset}} & \multicolumn{3}{c|}{\textbf{Cross-dataset}} & \multicolumn{2}{c|}{\textbf{DD-VQA}} \\
\textbf{Method} & FF++ c23 & FF++ c40 & CDF & DFDC & WDF & c23 & c40 \\ \hline
\ewf{}\textsubscript{SwinV2}       & 99.1 (0.97) & 95.0 (0.84) & 95.4 (0.74) & 88.3 (0.76) & 87.7 (0.67) & 95.0 (0.89) & 88.0 (0.71) \\
\ewf{}\textsubscript{XceptionNet}  & 99.3 (0.98) & 96.1 (0.88) & 99.0 (0.90) & 93.0 (0.85) & 95.9 (0.85) & 98.8 (0.97) & 89.3 (0.74) \\
\ewf{}\textsubscript{EfficientNet} & 98.9 (0.97) & 99.4 (0.98) & 98.5 (0.74) & 94.2 (0.88) & 96.5 (0.93) & 96.7 (0.93) & 95.5 (0.89) \\ \hline
\ewf{}\textsubscript{LAA-Net} (SBI)  & 82.0 (0.65) & -- & 79.3 (0.60) & 84.0 (0.43) & 67.1 (0.34) & -- & -- \\
\ewf{}\textsubscript{LAA-Net} (FF++) & 60.0 (0.28) & -- & 63.7 (0.29) & 47.7 (0.10) & 37.8 (0.10) & -- & -- \\ \hline
\end{tabular}}
\end{table*}

\section{Region-level results}
\label{sec:region-level_c40}

Tables~\ref{tab:region_ddvqa_c23} and~\ref{tab:region_ddvqa_c40} report the
region-level results at c23 and c40, which are summarized in the main paper.
At c40, as at c23, every \ewf{} variant answers every question type with more accurate verdicts than Qwen3-VL-4B trained on the same data (0.79--0.90 vs.\ at most 0.78), with \ewf{}\textsubscript{EfficientNet} best on all six. In CIDEr, the best \ewf{} variant outperforms both Qwen3-VL-4B models on every question type (\eg, 4.40 vs.\ 3.79 for the skin and 4.05 vs.\ 3.34 for the nose), although single variants fall below Qwen3-VL-4B (DD-VQA) for the eyes and the mouth. Without forensic tokens, adding FF++ before DD-VQA lowers both accuracy and CIDEr on every question type, consistent with the overall results in Table~\ref{tab:training_ddvqa_c40}.

\section{Decision consistency with the detector}
\label{sec:supp_consistency}

Table~\ref{tab:supp_consistency} reports how often each \ewf{} variant and the detector it integrates give the same verdict on the same test samples. For the three detectors trained on FF++ forgeries, the verdicts coincide on 95.0--99.4\% of the FF++ videos and 87.7--99.0\% of the videos of the unseen datasets (Cohen's $\kappa$ 0.67--0.98), and the video-level scores are strongly correlated (Spearman's $\rho$ 0.86--0.99). \ewf{} also follows the detector when it errs: on the unseen datasets, it gives the same wrong verdict for 82--98\% of the videos that the detector misclassifies, while it is wrong on 0--13\% of those that the detector classifies correctly. The same holds after explanation fine-tuning: on DD-VQA, the verdict in the generated answer agrees with the detector on 88.0--98.8\% of the test images ($\kappa$ 0.71--0.97), and matches the thresholded score $s(x,q)$ on at least 99.5\% of the questions. For LAA-Net, the SBI-aligned variant agrees less with its detector at the 0.5 threshold (67.1--84.0\%), at which LAA-Net alone is poorly calibrated, while their scores remain correlated ($\rho$ 0.72--0.91). The FF++-aligned variant largely departs from the detector's decisions ($\kappa$ 0.10--0.29, $\rho$ 0.36--0.64), in line with the loss of generalization reported in the main paper. These results concern the decisions of \ewf{}; they do not establish that the wording of an explanation reflects the detector's evidence.

\section{Qualitative examples}
\label{sec:supp_qualitative}
\providecommand{\ewf}{EwF}
\providecommand{\qualdir}{figures/qualitative}
\begin{figure*}[t]
\centering
\footnotesize
\begin{tabular}{@{}m{0.13\textwidth}@{\hspace{0.012\textwidth}}m{0.85\textwidth}@{}}
\includegraphics[width=\linewidth]{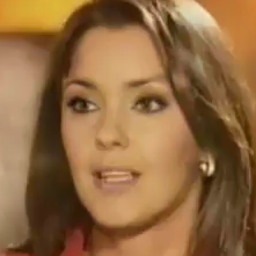} &
\textbf{Question:} Does the image looks real/fake? \hfill (image label: \textbf{fake}) \newline
\textbf{Reference:} \emph{The image looks fake. It looks like an ai-generated image although with human appearance but with unrealistic texture or lighting. The person's nose looks a bit fake because the nose is placed at the wrong place on the face. The person's skin looks a bit fake because there are stains/flaws on the person's face there are areas looking unnaturally brighter there area of looking unnaturally darker the person's mouth looks a bit fake because the mouth looks too rigid to convey expressions.} \newline
\textbf{\ewf{}\textsubscript{SwinV2}} \textcolor{green!50!black}{$\checkmark$}\textbf{:} The image looks fake. It looks like a computer-generated artist styled image for animation, carton or films. The person's mouth looks a bit fake because the mouth is misaligned with other facial features the person's eyes look a bit fake because the eyes look unnaturally asymmetric. The person's nose looks a bit fake because the nose lookes curved. \newline
\textbf{Qwen3-VL-4B (DD-VQA)} \textcolor{red!70!black}{$\times$}\textbf{:} The image looks real. Proper hair and hairstyles. The person has smooth skin, arched eyebrows, round eyes, straight nose, full mouth. \\[4pt]\hline\noalign{\vskip 4pt}
\includegraphics[width=\linewidth]{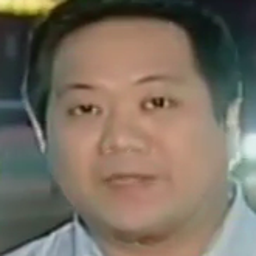} &
\textbf{Question:} Does the person's mouth look real/fake? \hfill (image label: \textbf{fake}) \newline
\textbf{Reference:} \emph{The person's mouth looks fake. The person's mouth looks very fake because overly small mouth.} \newline
\textbf{\ewf{}\textsubscript{SwinV2}} \textcolor{green!50!black}{$\checkmark$}\textbf{:} The person's mouth looks fake. The person's mouth looks a bit fake because the mouth is misaligned with other facial features. \newline
\textbf{Qwen3-VL-4B (DD-VQA)} \textcolor{red!70!black}{$\times$}\textbf{:} The person's mouth looks real. The person has naturally full mouth. \\[4pt]\hline\noalign{\vskip 4pt}
\includegraphics[width=\linewidth]{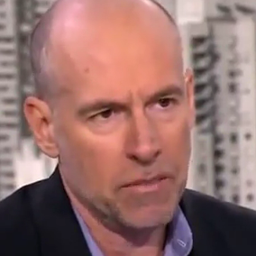} &
\textbf{Question:} Does the image looks real/fake? \hfill (image label: \textbf{real}) \newline
\textbf{Reference:} \emph{The image looks real. Matched beard or mustache. The person has smooth illuminated wrinked even skin, straight thin sparse eyebrows, oval small sparking eyes, straight nose, pouty mouth.} \newline
\textbf{\ewf{}\textsubscript{SwinV2}} \textcolor{green!50!black}{$\checkmark$}\textbf{:} The image looks real. Proper hair and hairstyles. The person has smooth skin, arched eyebrows, round eyes, straight nose, full mouth. \newline
\textbf{Qwen3-VL-4B (DD-VQA)} \textcolor{green!50!black}{$\checkmark$}\textbf{:} The image looks real. The person have complete face feature. Those features are in good shape, size, and positioning. The person has smooth skin, arched eyebrows, round eyes, straight nose, full mouth. \\[4pt]\hline\noalign{\vskip 4pt}
\includegraphics[width=\linewidth]{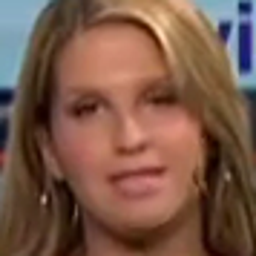} &
\textbf{Question:} Does the person's skin look real/fake? \hfill (image label: \textbf{fake}) \newline
\textbf{Reference:} \emph{The person's skin looks fake. The person's skin looks a bit fake because there area of looking unnaturally darker.} \newline
\textbf{\ewf{}\textsubscript{SwinV2}} \textcolor{green!50!black}{$\checkmark$}\textbf{:} The person's skin looks fake. The person's skin looks a bit fake because the skin is overly smooth and lacks of details, such as skin texture. There are stains/flaws on the person's face. There are areas looking unnaturally brighter. There area of looking unnaturally darker. \newline
\textbf{Qwen3-VL-4B (DD-VQA)} \textcolor{green!50!black}{$\checkmark$}\textbf{:} The person's skin looks fake. The person's skin looks a bit fake because the skin is overly smooth and lacks of details, such as skin texture. There are stains/flaws on the person's face. There are areas looking unnaturally brighter. There area of looking unnaturally darker. \\[4pt]\hline\noalign{\vskip 4pt}
\includegraphics[width=\linewidth]{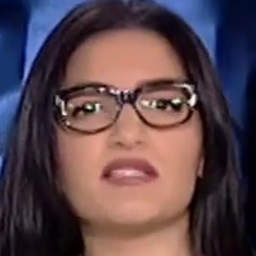} &
\textbf{Question:} Do the person's eyebrows look real/fake? \hfill (image label: \textbf{fake}) \newline
\textbf{Reference:} \emph{The person's eyebrows look fake. The person's eyebrows look a bit fake because the eyebrows look asymmetric.} \newline
\textbf{\ewf{}\textsubscript{SwinV2}} \textcolor{red!70!black}{$\times$}\textbf{:} The person's eyebrows look real. The person has naturally arched eyebrows. \newline
\textbf{Qwen3-VL-4B (DD-VQA)} \textcolor{red!70!black}{$\times$}\textbf{:} The person's eyebrows look real. The person has naturally arched eyebrows. \\
\end{tabular}
\caption{Qualitative examples on DD-VQA (c23): answers of \ewf{}\textsubscript{SwinV2} (trained on FF++ \& DD-VQA) and Qwen3-VL-4B (trained on DD-VQA) to the same question, with one of the reference answers. $\checkmark$/$\times$: the verdict in the answer agrees/disagrees with the label of the image. The examples were selected from random candidates to show cases in which only \ewf{} gives the correct verdict (first two rows), both models do (third and fourth rows) and both fail (last row). Questions, references and answers are reproduced verbatim, including the spelling of the dataset.}
\label{fig:supp_qualitative_c23}
\end{figure*}

Figure~\ref{fig:supp_qualitative_c23} shows answers of \ewf{}\textsubscript{SwinV2} and Qwen3-VL-4B to image- and region-level questions. For Qwen3-VL-4B, we used the model trained on DD-VQA only, the better of its two variants in explanation. In the first two examples, only \ewf{} recognizes the manipulation, both for the whole face, with an explanation that names facial regions also mentioned in the reference (the nose and the mouth), and for the mouth alone. For real faces and for some manipulated regions, both models give correct and similar answers, while in the last example both models miss a manipulation of the eyebrows. As discussed in the main paper, text metrics measure agreement with the reference answers, not whether an explanation reflects the evidence of the detector.

\end{document}